\documentclass[sigconf]{aamas} 

\usepackage{balance} 

\setcopyright{ifaamas}
\acmConference[AAMAS '24]{Proc.\@ of the 23rd International Conference
on Autonomous Agents and Multiagent Systems (AAMAS 2024)}{May 6 -- 10, 2024}
{Auckland, New Zealand}{N.~Alechina, V.~Dignum, M.~Dastani, J.S.~Sichman (eds.)}
\copyrightyear{2024}
\acmYear{2024}
\acmDOI{}
\acmPrice{}
\acmISBN{}

\acmSubmissionID{271}

\title[AAMAS-2024 Formatting Instructions]{Minimax Exploiter: A Data Efficient Approach for Competitive Self-Play}

\author{Daniel Bairamian}
\affiliation{
  \institution{McGill University}
  \city{Montreal}
  \country{Canada}}
\email{daniel.bairamian@mail.mcgill.ca}

\author{Philippe Marcotte}
\affiliation{
  \institution{Ubisoft Montreal}
  \city{Montreal}
  \country{Canada}}
\email{philippe.marcotte@ubisoft.com}

\author{Joshua Romoff}
\affiliation{
  \institution{Ubisoft Montreal}
  \city{Montreal}
  \country{Canada}}
\email{joshua.romoff@ubisoft.com}

\author{Gabriel Robert}
\affiliation{
  \institution{Ubisoft Montreal}
  \city{Montreal}
  \country{Canada}}
\email{gabriel.robert@ubisoft.com}

\author{Derek Nowrouzezahrai}
\affiliation{
  \institution{McGill University}
  \city{Montreal}
  \country{Canada}}
\email{derek@cim.mcgill.ca}

\begin{abstract}
Recent advances in Competitive Self-Play (CSP) have achieved, or even surpassed, human level performance in complex game environments such as \textsc{Dota 2} and \textsc{StarCraft II} using Distributed Multi-Agent Reinforcement Learning (MARL). 
One core component of these methods relies on creating a pool of learning agents -- consisting of the Main Agent, past versions of this agent, and Exploiter Agents -- where Exploiter Agents learn counter-strategies to the Main Agents. A key drawback of these approaches is the large computational cost and physical time that is required to train the system, making them impractical to deploy in highly iterative real-life settings such as video game productions. In this paper, we propose the Minimax Exploiter, a game theoretic approach to exploiting Main Agents that leverages knowledge of its opponents, leading to significant increases in data efficiency. We validate our approach in a diversity of settings, including simple turn based games, the arcade learning environment, and \textsc{For Honor}, a modern video game. The Minimax Exploiter consistently outperforms strong baselines, demonstrating improved stability and data efficiency, leading to a robust CSP-MARL method that is both flexible and easy to deploy.
\end{abstract}


\newcommand{\BibTeX}{\rm B\kern-.05em{\sc i\kern-.025em b}\kern-.08em\TeX}

\begin{document}


\pagestyle{fancy}
\fancyhead{}


\maketitle 


\section{Introduction}
\label{introduction}

Reinforcement learning (RL) has demonstrated its utility in reliably solving tasks in static environments, such as with Atari games \cite{AtariDRL} and diverse control problems \cite{OpenAIGym}. In \textit{competitive environments}, however, choosing an adequate adversary for the RL agent becomes a design problem with important challenges in and of itself. Notably, choosing too hard of an opponent can limit the ability of an agent to learn, while choosing too easy of an opponent can lead to sub-optimal learned agent policies. 
Recent advances in \textsc{Starcraft II} \cite{alphastar}, \textsc{Go} \cite{Go}, \textsc{Dota 2} \cite{Dota2}, \textsc{Gran Turismo} \cite{sophie}, \textsc{Chess} and \textsc{Shogi} \cite{ChessAndShogi} 
have successfully demonstrated human-level performance by leveraging Competitive Self-Play (CSP) \cite{CSP}: having an agent play against itself provides a balance between the two aforementioned extremes. CSP trains RL agents in a league setup, where a matchmaking algorithm pairs different agents against each other based on a performance metric, such as win-rate or Elo rating \cite{ELORating}. 

\begin{figure}[t!]
\begin{center}
\centerline{\includegraphics[width=\columnwidth]{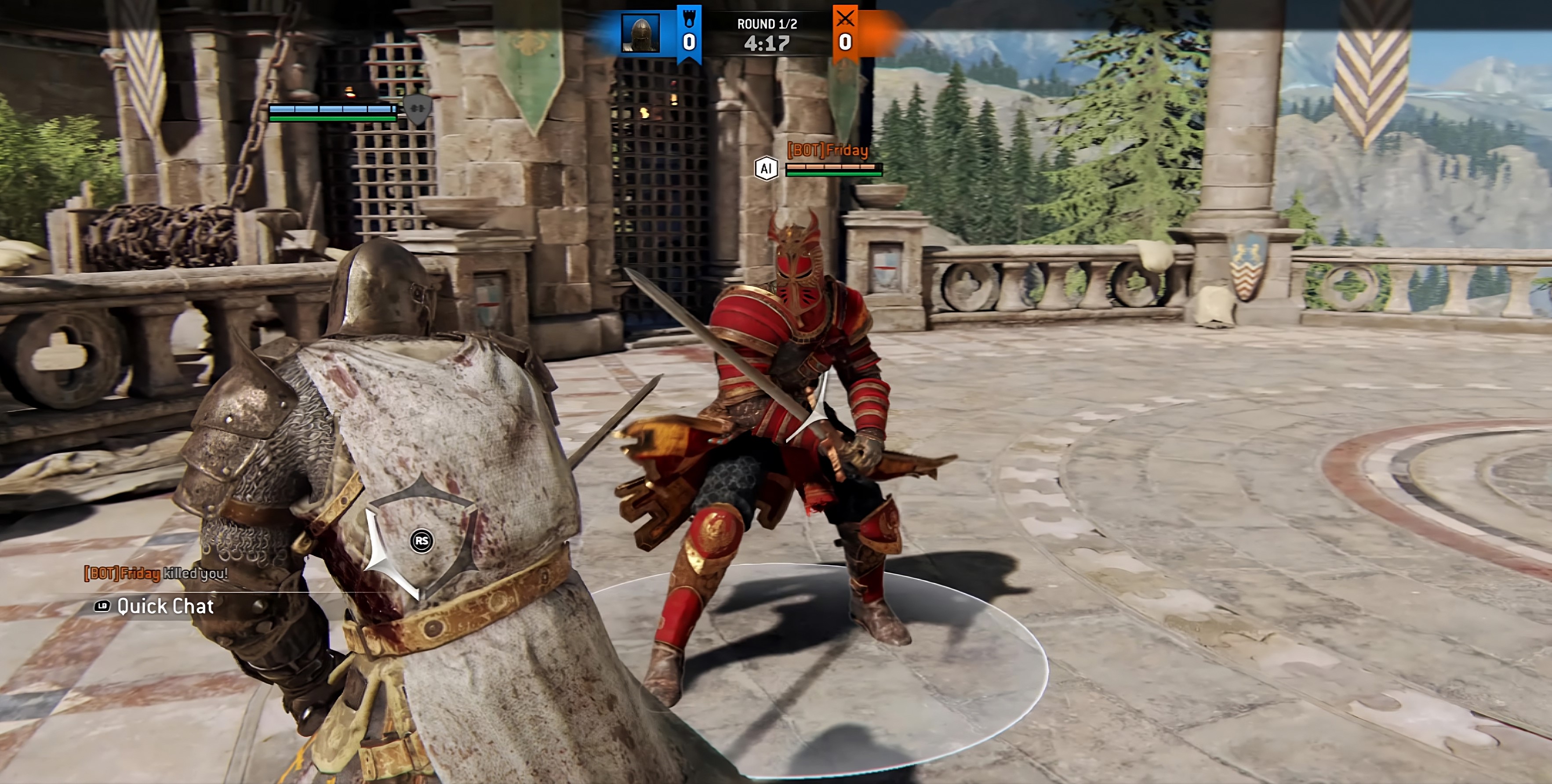}}
\vspace{-5pt}
\caption{Our Deep RL agent in the \textsc{For Honor} game: screenshot from our testing environment, not from live private player views.}
\label{figure:fh_screenshot}
\end{center}
\vskip -0.4in
\end{figure}

Two main archetypes exist in the league framework: the first is a Main Agent, whose role is to learn a robust strategy by competing against the league's opponent pool; the second is a Main Exploiter, whose role is to train specifically against a single instance of the Main Agent, with the goal of learning a counter strategy. Unfortunately, this league training paradigm -- often referred to as Competitive Self-Play Multi Agent Reinforcement Learning (CSP-MARL) -- is extremely expensive to run, despite recent efforts to improve its efficiency \cite{Tleague, TStarBotX}. For example, \textsc{TLeague} \cite{Tleague} trains a $20$ million parameter model in $57$ days with roughly $1.3 \times 10^4$ CPUs and $144$ GPUs, compared to \textsc{Alphastar}'s \cite{alphastar} $139$ million parameter model in $44$ days with $5 \times 10^5$ CPUs and $3\times 10^3$ GPUs, all while achieving comparable results. As training prolongs, it becomes increasingly challenging for exploiters to find counter strategies, and so convergence times become a training bottleneck \cite{Tleague}. 

Crucially, long training times can prevent CSP-MARL from being consistently used in highly iterative workflows such as video game development, where the environment dynamics from the RL agents point of view can change on a daily basis. A constantly changing environment makes the multi-week (or even multi-month) training times completely infeasible for many use cases in the development cycle, such as balancing different classes, or testing design changes \cite{seedTesting}.


To this end, we propose an alternative to the standard Exploiter archetype, we call the Minimax Exploiter, that utilises a game theoretic inspired reward function that aims to minimize the maximum Q-value of its opponent at every step. We show empirically that our Exploiter converges faster than standard exploiters in a variety of settings. We first evaluate the efficiency of the Minimax Exploiter on \textsc{Tic-Tac-Toe}, \textsc{Connect 4} and \textsc{Atari Boxing}, before stress testing our approach on a modern AAA \footnote{AAA is an informal rating given to games with high budgets.} video game environment, For Honor, where we deploy a complete league training setup to evaluate its efficiency.

\section{Background and Related Work}
\label{background}

Two main schools of thought exist in recent self-play model free RL works. The first, popularized by \textsc{StarCraft} and \textsc{Dota} \cite{alphastar, Dota2}, trains an agent in a league setup against a population of past versions of itself while incorporating  human data to bootstrap training. In contrast to the league training methodology, game theoretic approaches such as \textsc{DeepNash}, a recent \textsc{Stratego}-playing agent  \cite{Stratego} (which builds off prior work \cite{NeuRD}) introduce a Regularized Nash Dynamics algorithm (R-NaD) that seeks a Nash Equilibrium via policy regularization \cite{PoinCarre}.
Motivated by the former work's ability to solve complex modern video games, we build atop league training in our approach.  


When training an agent in a league setup, a popular architecture used by \textsc{Alphastar} \cite{alphastar} trains three unique agent archetypes. The first archetype -- the Main Agent -- is trained using prioritized fictitious self-play (PFSP), sampling opponents for the Main Agent from an opponent pool that mainly consists of past versions of the agent itself (and proportional to their associated win-rates). The motivation is to promote sampling the toughest opponents to force the Main Agent to always improve. The second archetype -- the Main Exploiter -- is trained solely against the latest version of the Main Agent. While the Main Agent learns a diversity of strategies, the Main Exploiter seeks only to learn those strategies that defeat (or weaken, exploit) the Main Agent. Once converged, the Main Exploiter is added to the opponent pool used to train the next iteration of the Main Agent (again, using PFSP), promoting the development of strategies which correct any weaknesses found earlier by the Main Exploiter. The final archetype -- the League Exploiter -- is also trained using PFSP, but against the \textit{entire history} of the league, rather than just its latest iteration of the Main Agent. 
Just like the Main Exploiter, their goal is to find weaknesses, however rather than being targeted specifically to the Main Agent, they are targeted to the entire league.

Such frameworks generally consists of multiple different components, which work in tandem to assure a functioning league. The actors are responsible for interacting with the environment, generating experience data. The learners, are responsible for training the networks given the experience data. The model pool consolidates all different networks that are used in the league, and finally the league manager, coordinates the different modules together, as well as samples the opponents for the Main Agent, through its sub-module called the game manager. Tencent's \textsc{TLeague} framework \cite{Tleague, TStarBotX}, uses these modules, as well as additional ones, such as their hyperparameter manager and inference server, to further optimize this process.


In complex environments, e.g., \textsc{StarCraft}, imitation learning with expert human data is often used to bootstrap agent policy learning. In \textsc{Alphastar}, human data (roughly $10^6$ game replays) is first clustered by strategy type and encoded into a policy network $\pi_{\theta}(a_t|s_t, z)$ parameterized by a player's build order $z$, i.e., the first $20$ builds and units constructed by the player. This approach both initializes agents with a strong bootstrap model and promotes strategy diversity in the league.

\section{Motivating Example -- \textsc{Tic-Tac-Toe}}
\label{sec:motivating}
We motivate our approach with a simple zero-sum game, \textsc{Tic-Tac-Toe}. The rules of the game are simple, players take alternating turns placing either an X (player $1$) or and O (player $2$) on a $3\times 3$ grid. A game ends when either one player wins by placing three of its symbols in sequence (horizontally, vertically, or diagonally), or a draw is reached.


Consider playing against an optimal \textsc{Tic-Tac-Toe} player, which can be easily and efficiently obtained by the Minimax algorithm given the game's small game tree. Training an RL agent against the optimal Minimax player is realizable without any changes to classic RL algorithms: with fewer than $3^9$ game states\footnote{A $3\times 3$ grid of \{X, O, empty\} cells, less invalid configurations.}, nine possible actions, and a simple reward of $+1$ if you win, $-1$ if you lose, and $0$ if you draw.

\begin{figure}[t!]
\vskip 0.2in
\begin{center}
\centerline{\includegraphics[width=\columnwidth]{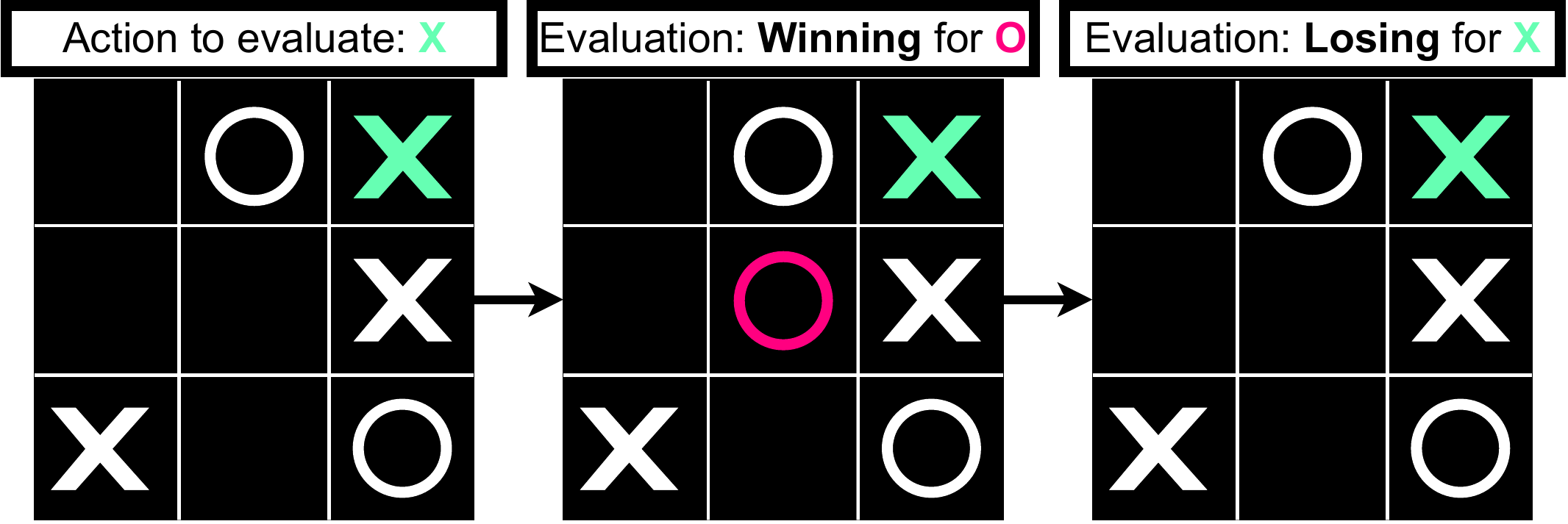}}
\vspace{-5pt}
\caption{\label{figure:TicTacToeEval}\textsc{Tic-Tac-Toe} expert negative evaluation is propagated back to the agent and used as a reward proxy.}
\end{center}
\vskip -0.2in
\end{figure}

This simple reward function is, however, sub-optimal. For example,
the agent would only know that their initial move is losing upon episode termination, with this information propagating back to the original action that lead to the loss. Noting that the Minimax evaluation of the position -- being exact -- altogether replaces the need for a reward function, we could use the Minimax evaluation directly to estimate the value of an action. Figure~\ref{figure:TicTacToeEval} diagrams this idea of using the expert evaluation of a game as a reward proxy.

In complex games with large (or continuous) state and/or action spaces, a Minimax evaluation is not tractable nor is it common to have access to any such oracle evaluation. In the case of self-play, however, another viable option arises: it is usually both cheap and meaningful to evaluate an action \textit{from the perspective of your opponent}. Specifically, when training an Exploiter (tasked with defeating a single, frozen instance of the Main Agent), it seems natural to rely on the Main Agent's evaluation of the position to aid in the Exploiter's training. We delve deeper into this idea before outlining our approach, below.

\section{Approach}
\label{Approach}
Our approach aims at speeding up the training of Exploiter Agents in competitive self-play, while preserving the optimal policy of the Main Agent.
We treat two player zero-sum games, proposing a methodology to accelerate Exploiter training when it is possible to gain access to the Main Agent's action evaluation, such as through its Q-function, Value function, or any other appropriate proxy. Our core principle consists of propagating the negative evaluation of the Main Agent's state to the Exploiter's state-action pair that leads to that state (see Figure~\ref{figure:TicTacToeEval}). While any RL algorithm could, in theory, eventually converge to the same result, we demonstrate that providing the Main Agent's evaluation can accelerate the convergence process. We first detail the theoretical foundation of our approach.  

\subsection{Theoretical Foundation}
\label{sec:theoreticalfoundation}
We explore the theoretical foundations of our approach, starting by assuming that we are in a zero sum game, i.e., that $R^i_t = -R^{j}_t$, where $t$ is the timestep, $i$ denotes the first player and $j$ the second. Thus, a positive reward for one agent will result in the equivalent negative reward for the other.

It follows that the returns of both agents also reflect this zero-sum property, which -- over a fixed horizon -- we now show to be, without loss of generality:
\begin{equation}
\label{eq:equivalency}
   G_t^i =\sum_{t=0}^T  \gamma^t R^i_t = - \sum_{t=0}^T  \gamma^t R^{j}_t = -G_t^{j}~,
\end{equation}
where $\gamma$ is the discount factor. Given the equivalency in Equation~\ref{eq:equivalency}, we can rewrite $Q^i$, the Q function for agent $i$,  $Q: \mathcal{S} \times \mathcal{A}\rightarrow \mathbb R$, i.e., the expected sum of discounted rewards from a given state and action, as a function of $Q^j$, the Q function for agent $j$:
\begin{align}
&Q^i (s, a) = \mathbb E^{\pi_i, \pi_{j} } \left [ G^i_t \vert s_t=s, a_t=a \right ] \nonumber \\
&= \mathbb E^{\pi_i, \pi_{j} } \!\!\left [ R^i_t + G^i_{t+1} \vert s_t=s, a_t=a \right ] \nonumber \\
&= \mathbb E^{\pi_i, \pi_{j} } \!\!\left [ R^i_t - G^{j}_{t+1} \vert s_t=s, a_t=a \right ] \nonumber \\
&= \mathbb E^{\pi_i, \pi_{j} } \!\!\left [ R^i_t - Q^{j} (s_{t+1}, a_{t+1}) \vert s_t=s, a_t=a \right ]\! \nonumber \\
&= \mathbb E^{\pi_i, \pi_{j} } \!\!\left [ R^i_t - V^{j} (s_{t+1}) \vert s_t=s, a_t=a \right ]\!\label{eq:q_equivalence},
\end{align}
where on the last line we replace the $Q^j$ with $V^j$, the value function $V: \mathcal{S} \rightarrow \mathbb{R}$, i.e., the expected discounted rewards from a given state. We also note that both Q and V are defined as the expectation over \textit{both} agents' policies, $\pi_i$ and $\pi_{j}$, where $\pi: \mathcal{S} \rightarrow dist(\mathcal{A})$.  

For Equation~\ref{eq:q_equivalence} to hold and be usable in practice, the opponents value function would have to accurately represent the current policy, $\pi_i$. We argue, however, that this will almost never be the case in practice. This is particularly true when training against fixed opponents that are not able to adapt to the current agent's strategy. We note that this is precisely the case for Exploiter Agents that use a fixed Main Agent to exploit. In the following sections, we explore an alternative solution for using the opponent's evaluation.

\subsection{The Minimax Reward}
\label{sec:minimaxreward}

In practice, since the Main Agent's value function will be inaccurate, we propose to use a reward function that mixes both the environment reward and the opponent's value function. Specifically, assuming the opponent is acting greedily with respect to its value function, we have $V^j(s_{t+1}) = \max_a Q^j(s_{t+1}, a)$, and therefore use the following reward, which we term the \textit{Minimax} reward:
\begin{equation}
R^i_{\textit{minimax}}(s_{t}, a_{t}) = R^i(s_{t}, a_{t}) - \alpha \gamma (1-d) \max_a Q^j(s_{t+1}, a),
\label{eq:reward_exploiter}
\end{equation}
where $\alpha \in \left [0, 1 \right ]$ is a coefficient modulating the opponent's signal, $d \in \{0, 1\}$ is the done signal, and $\max_a Q^j(s_{t+1}, a)$ is the maximum opponent's evaluation at the next state.  We note that the choice in $\alpha$ is important, since setting it too high will potentially result in the agent focusing too much on the opponents value function, which in our case is only an approximation.


Moreover, to reduce the potential for reward hacking via finding a cycle of infinite positive rewards, we ensure that the additional reward term based on the opponents value function, provided only at non-terminal states, is at most zero. Specifically, in sparse reward environments, the Main Agent's value function will approximately (barring function approximation errors), lie within the bounds of the reward function $R \in \left [ R_{\textit{min}}, R_{\textit{max}} \right ]$. In this setting, we simply shift the additional reward of our Minimax Exploiter by $-\vert R_{\textit{min}} \rvert$ such that $-V^j(s_{t+1}) -\vert R_{\textit{min}} \rvert  \leq 0$, which can be seen by the following:
\begin{align}
\label{eq:reward_shifting}
- V^j(s_{t+1}) &\leq  -R_{\textit{min}} \nonumber \\
-V^j(s_{t+1}) - \lvert R_{\textit{min}} \rvert &\leq -R_{\textit{min}} - \lvert R_{\textit{min}} \rvert \nonumber \\
-V^j(s_{t+1}) - \lvert R_{\textit{min}} \rvert &\leq 0, 
 \end{align}
 where we note that when $R_{\textit{min}}$ is negative, the right hand side of the last equation is $0$, and when it's positive, the right hand side of the equation will be negative.

In cases where the underlying reward function is not sparse, or when knowledge of the minimum reward is not available, we suggest simply tracking the minimum value, $\min_s V^j(s)$ while the Main Agent was training and shift the reward negatively by that amount. 

We note that using the \textit{Minimax} reward, with or without this shift, does not theoretically guarantee that the original optimal policy is preserved. However, this modified reward is only used on Exploiter Agent's, to speed up their convergence speed, and not on the Main Agent. 

\begin{figure}[t!]
\vskip 0.2in
\begin{center}
\centerline{\includegraphics[width=1.0\columnwidth]{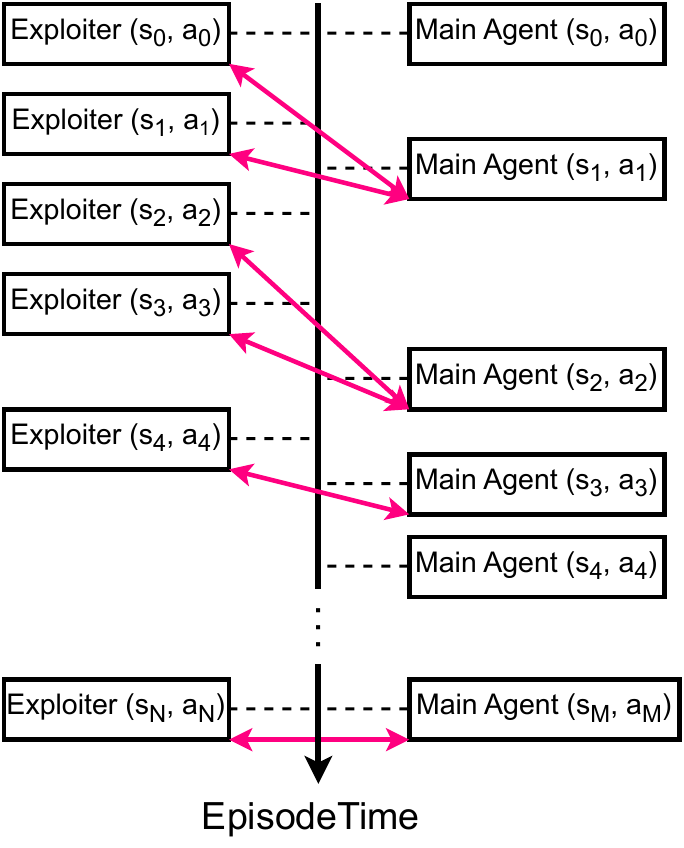}}
\caption{\textsc{For Honor} Exploiter and Main Agent asynchronous state-action pairings based on chronological order, as occurred within an episode,  denoted by the double red arrows.}
\label{figure:Tuples_FH}
\end{center}
\vskip -0.4in
\end{figure}

\paragraph{Note on Reward Pairing:}
In standard turn-based games, the state in $s_{t+1}$ from $V^j(s_{t+1})$ in Equation~\ref{eq:reward_exploiter} is simply the next turn of the opponent, however -- in simultaneous games like \textsc{For Honor}, where both agents act asynchronously -- we simply pair state-action pairs based on their chronological occurrence (see Figure~\ref{figure:Tuples_FH}). We note that both players are not guaranteed to have the same number of state-action pairs throughout an episode, as one player may act multiple times before allowing its opponent to act. This is common in, e.g., fighting games where stunning/disabling your opponent is a foundational principle of effective gameplay. In these settings, we will have non-unique pairings of some state-action pairs.

\section{Experiments}
\label{experiments}
We evaluate our approach on games of increasing complexity, starting with the simple turn based games \textsc{Tic-Tac-Toe} and \textsc{Connect 4}, then \textsc{Atari Boxing} and finally \textsc{For Honor} a AAA video game. Since the Minimax reward requires an extra network evaluation, it would be unfair to baseline approaches to judge the training performance over a fixed number of network updates or environment steps. Therefore, all experiments are evaluated over a predetermined wall-clock training time. Note that all of our benchmarks are Deep Q-Network (DQN)\cite{AtariDRL} agents utilising a double Q-function \cite{DoubleQLearning}, that we simply refer to as the DQN agent.  We plan on releasing (upon acceptance) \textsc{Tic-Tac-Toe}, \textsc{Connect 4} and modified \textsc{Atari Boxing} with two player support source code for reproducibility purposes.

\subsection{Turn Based Games}
\label{sec:turn}
We evaluate our framework on \textsc{Tic-Tac-Toe} and \textsc{Connect 4}, each two player zero-sum turn based games. When the agent plays a move, the environment will respond by picking the best move according to a Minimax algorithm, which we use as our proxy to a valuation function. For \textsc{Tic-Tac-Toe}, we allow the Minimax algorithm to run to completion -- representing the optimal policy. For \textsc{Connect 4}, rather than letting the Minimax algorithm run to completion, we limit its search depth to only three moves; this means that the policy is effectively random, unless there is an opportunity to win or prevent a loss within a three move window. The motivation here is to have a scenario where we know that our opponent's action evaluation is imperfect. In both these environments, the reward structure is $+1$ when the agent wins, $-1$ when the agent loses, and $0$ otherwise.

\begin{figure*}[t!]
\centering
\includegraphics[width=\columnwidth] {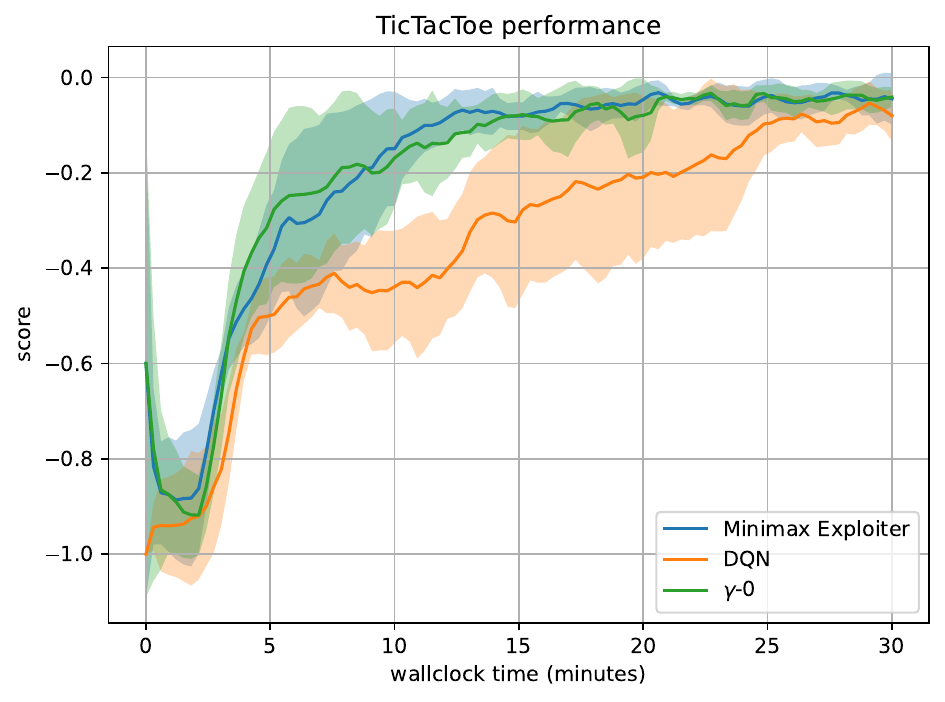} \hfill
\includegraphics[width=\columnwidth]{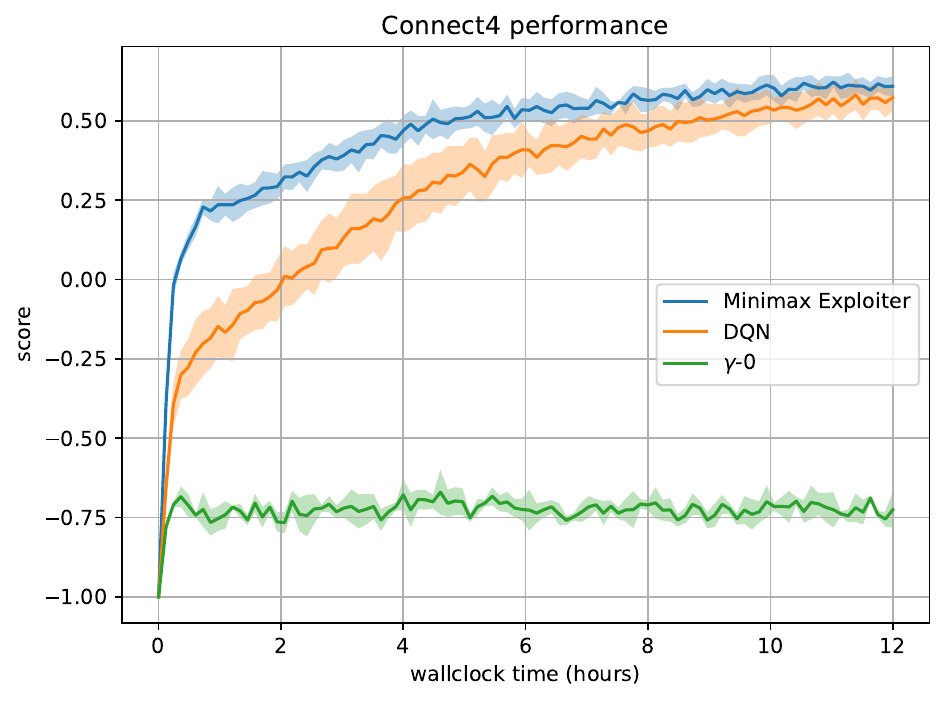} 
\caption{\textbf{left:} \textsc{Tic-Tac-Toe}, \textbf{right:} \textsc{Connect 4}, training performance of the Minimax Exploiter, DQN agent, and $\gamma$-0 agent, averaged over five seeds.The Minimax Exploiter is significantly more efficient than the DQN agent in both settings, the $\gamma$-0 agent is on par with the Minimax Exploiter in \textsc{Tic-Tac-Toe}, but unable to learn in \textsc{Connect 4}.}
\label{figure:all_results}
\end{figure*}

\paragraph{\textsc{Tic-Tac-Toe} Results:}
We train a DQN agent as well as a Minimax Exploiter agent against a Minimax opponent, with a Minimax Exploiter $\alpha = 0.1$. The agents train with an $\epsilon$-greedy exploration of $0.01$ and we evaluate performance over $30$ minutes of training, averaged over five seeds, which we illustrate in Figure~\ref{figure:all_results}. We train an additional agent (also in Figure~\ref{figure:all_results}), which we call $\gamma$-0: this agent is a Minimax Exploiter agent with a discount factor $\gamma = 0$ and a Minimax Exploiter $\alpha = 1$, effectively cloning the inverse Minimax of the opponent as its Q-function. Note that here we are referring to the $\gamma$ of the Q-learning TD update, and not the $\gamma$ from \eqref{eq:reward_exploiter}, which is always set to the default value of $0.995$. Each agent's Q-function is parameterized as a fully connected neural network with two hidden layers, each with a dimension of $64$. 

From our training performance, we see that the Minimax Exploiter and the $\gamma$-0 agent are significantly more efficient than a DQN agent. We can expect these gains to increase as environment complexity increases.

\paragraph{\textsc{Connect 4} Results:}
\textsc{Connect 4} is a slightly more complex experiment, where the training setup remains the same as in \textsc{Tic-Tac-Toe} except for having a Minimax Exploiter $\alpha = 0.01$ and the Q-function network parameterized with two hidden layers with dimensions of size $512$. We evaluate training over $12$ hours, rather than $30$ minutes, and average over three seeds rather than five. Figure~\ref{figure:all_results} illustrates that the Minimax Exploiter continues to converge much faster than the baseline DQN agent. Note that the $\gamma$-0 agent is unable to learn and collapses at a score of about $-0.75$. This example showcases that even with an imperfect evaluation, evidenced by the poor $\gamma$-0 performance, the Minimax Exploiter is still able to extract a useful signal to boost its training performance.


\subsection{\textsc{Atari Boxing}}
\label{sec:box}
We modify the \textsc{Atari Boxing} environment to end when one of the two fighters reaches ten points, as opposed to the default $100$, and use the \textsc{Pettingzoo} library \cite{PettingZoo} to run the game in two player mode. We first train a DQN agent against a random policy, then train a second agent against the first one. This second agent will serve as the environment opponent in our tests. We do this to emulate a league training setup that is several generation in, where bottlenecks of the Main Exploiter may become more apparent \cite{Tleague}. The reward function we use to train this agent remains sparse, with a $+1$ when reaching ten points, $-1$ if the opponent reaches ten points, and $0$ otherwise. 

The \textsc{Atari Boxing} experiments consist of training the Minimax Exploiter with $\alpha = 0.005$, a DQN agent that observed the sparse reward function which we call DQN-sparse, and another DQN agent that observed the dense reward function from the environment (i.e., which is $+1$ every time the agent successfully hits the opponent), which we call DQN-dense. The network sizes and exploration hyperparameters are the same as in \textsc{Connect 4}, and we evaluate training over three hours and averaged over five seeds. We note that we used the RAM observations rather than the traditional pixel observations.

\begin{figure}[t!]
\vskip 0.2in
\begin{center}
\centerline{\includegraphics[width=\columnwidth]{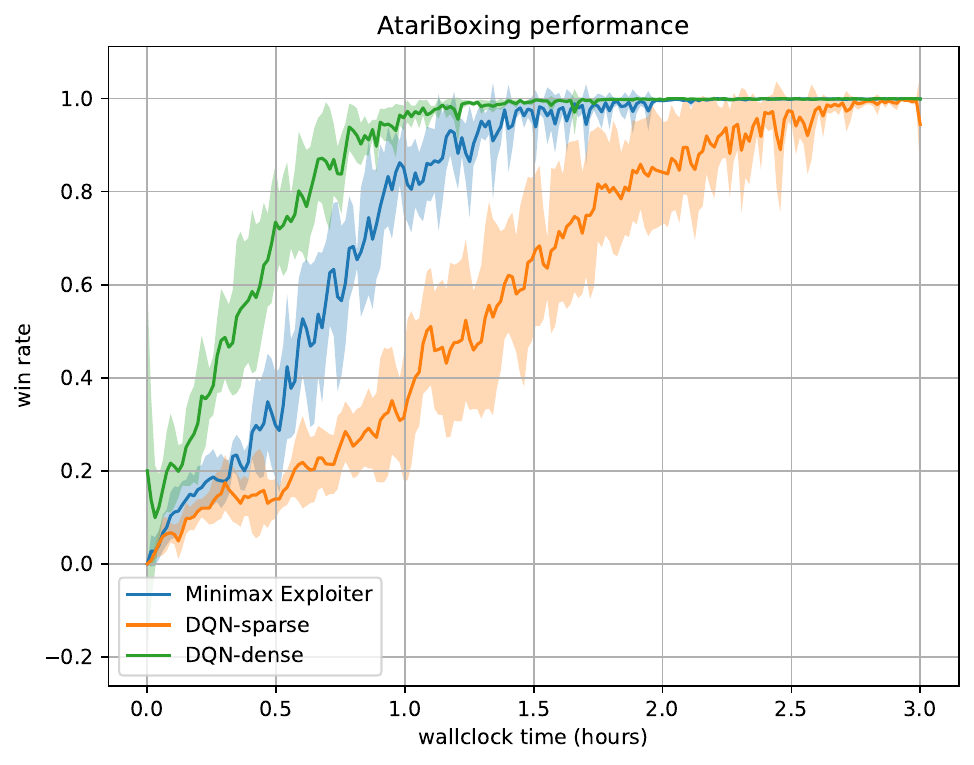}}
\vspace{-5pt}
\caption{\textsc{Atari Boxing} training performance over three hours of the Minimax Exploiter, DQN sparse agent, and DQN dense agent, averaged over five seeds. The Minimax Exploiter is significantly more efficient than the DQN sparse agent, however less efficient than the DQN dense agent.}
\label{figure:AtariBoxingperf}
\end{center}
\vskip -0.2in
\end{figure}

From the results in Figure~\ref{figure:AtariBoxingperf}, we can see that adding the Minimax reward greatly improves performance compared to the purely sparse reward function, and the default dense reward function from the environment (unsurprisingly) outperforms our Minimax Exploiter. Since there is little to no downside of getting hit in \textsc{Atari Boxing}, a dense reward function promoting aggressive behavior is expected to perform very well. However, this is a special case of simple reward engineering that leads to fast convergence, which may not be as trivial to determine in more complex environment \cite{RewardShaping}, as we showcase in the next set of experiments on \textsc{For Honor}. 

\subsection{For Honor}
\label{sec:forhonor}
Our final and most complex experiment scales and deploys a CSP-MARL setup to \textsc{For Honor}, a modern AAA action combat game with a variety of game modes. We consider the duel mode, a one-versus-one combat that ends when one of the two fighters lose all their health points. While the game has a variety of hero archetypes with different play styles, for simplicity, we exclusively treat Wardens in our training, which are melee combat knights with long swords.

The action space of the agents consist of $36$ discrete actions, comprising directional hits, dodges, blocks, among many other moves. The state space consists of a $160$ dimensional vector, which encodes information about each agent's health, stamina, speed, distance to the opponent, animation information, and other game specific information such as the opponent's stance. We provide the full details of our RL interface in the supplemental material.

The underlying framework we use is a simplified version of the one presented in \textsc{TLeague} \cite{Tleague}, with a few modifications to fit our settings. To reduce computation cost, we only train our league framework with two of the three mentioned archetypes, namely the Main Agent and Main Exploiter, omitting the League Exploiter. We also do not use any expert data as a starting point for our models. Finally, we define convergence for the Main Exploiter as having a win-rate of $85\%$ against its current frozen (non-training) Main Agent, and convergence for the Main Agent as having a win-rate of $85\%$ against all agents in the opponent pool. The Main Exploiter will only trigger a new generation if there exists a different Main Agent to exploit (i.e., the Main Agent has converged at least once since it last got paired with the Main Exploiter). The opponents of our Main Agent are sampled proportionally to their win-rates, with a $10\%$ chance of randomly sampling any opponent in the league.

We train by running four game clients on a single machine, each client running ten duels at double speed, leading to effectively $80$ duels simultaneously running on each machine. For our experiments, we only utilise a single machine to run the game clients. Of the four running game clients, two of them generate experience to train the Main Agent, while the other two clients generate experience to train the Main Exploiter. We run the league framework on a compute cluster with two GPUs.


We set the environment reward of our \textsc{For Honor} duels to yet another simple, sparse reward function, with a reward of $+10$ when the agent wins their duel, $-10$ if the agent loses, and $0$ otherwise.

Each agent utilizes a Deep Q-Network, with two fully connected hidden layers of dimension $512$ and an $\epsilon$-greedy exploration of $0.01$. We run four different league setups, with varying Main Exploiters. The first Main Exploiter observes the sparse reward function, which we call the Vanilla Exploiter. The second is our Minimax Exploiter with $\alpha = 0.01$. Finally, we add two additional exploiters with custom dense reward functions: the first providing a positive reward each time the agent hits their opponent -- which we call the Aggressive Exploiter -- and the second providing a negative reward each time the agent gets hit from their opponent -- which we call the Defensive Exploiter. Note that all the Main Agents observe the same sparse reward function, regardless of which type of Main Exploiter they are training against.

The first generation of the Main Agent is initialized to a model that has trained to convergence ($\geq 85\%$ win-rate) against the scripted AI that exist in the game, while the Main Exploiters start from a random initialization. After each generation, the Main Agent retains its current network, while the Main Exploiter does not. 

\begin{figure}[t!]
\vskip 0.2in
\begin{center}
\includegraphics[width=\columnwidth]{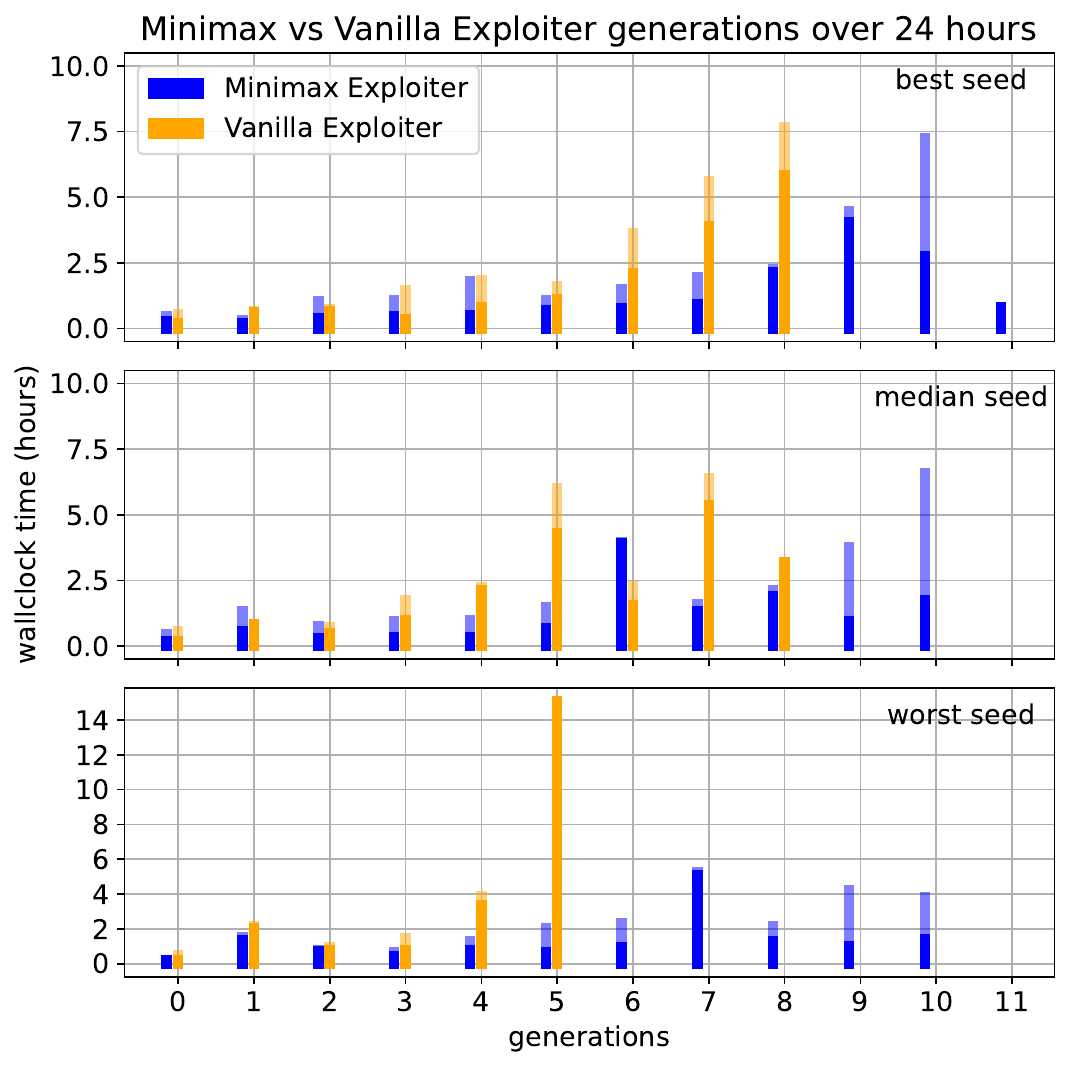}
\includegraphics[width=\columnwidth]{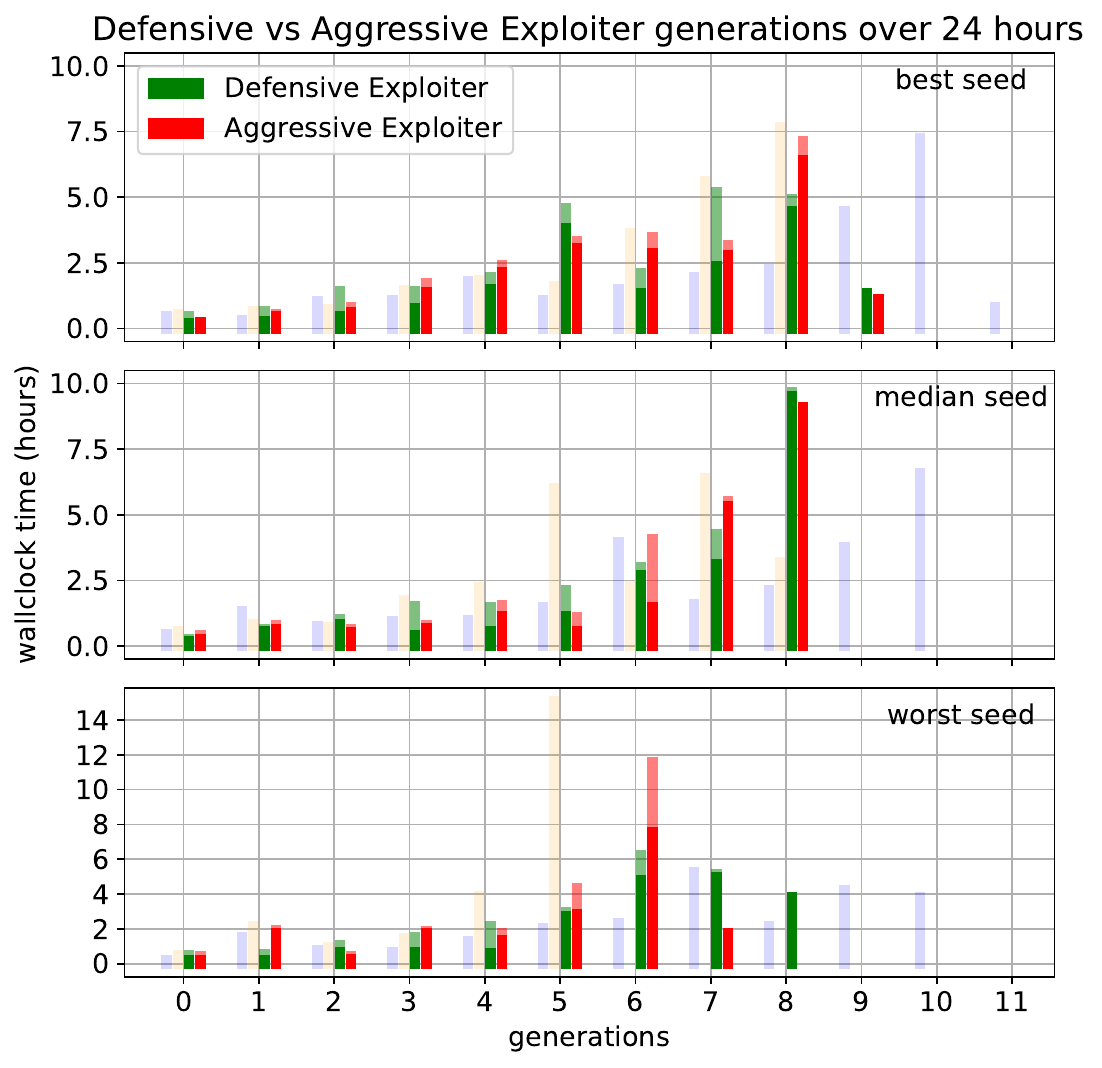}
\caption{Minimax vs Vanilla vs Defensive vs Aggressive Exploiter Performance over $24$ hours in \textsc{For Honor} with three different seeds. The Minimax Exploiter performs the best with $11$ converged generations at each seed. Both the Defensive and Aggressive Exploiters outperform the Vanilla exploiters with $9$ converged generations at their best seed.}
\label{figure:SmartExploiter_vs_Vanilla_vs_Defensive_vs_Aggressive_performance}
\end{center}
\vskip -0.3in
\end{figure}

\begin{figure}[ht!]
\vskip 0.2in
\begin{center}
\centerline{\includegraphics[width=\columnwidth]{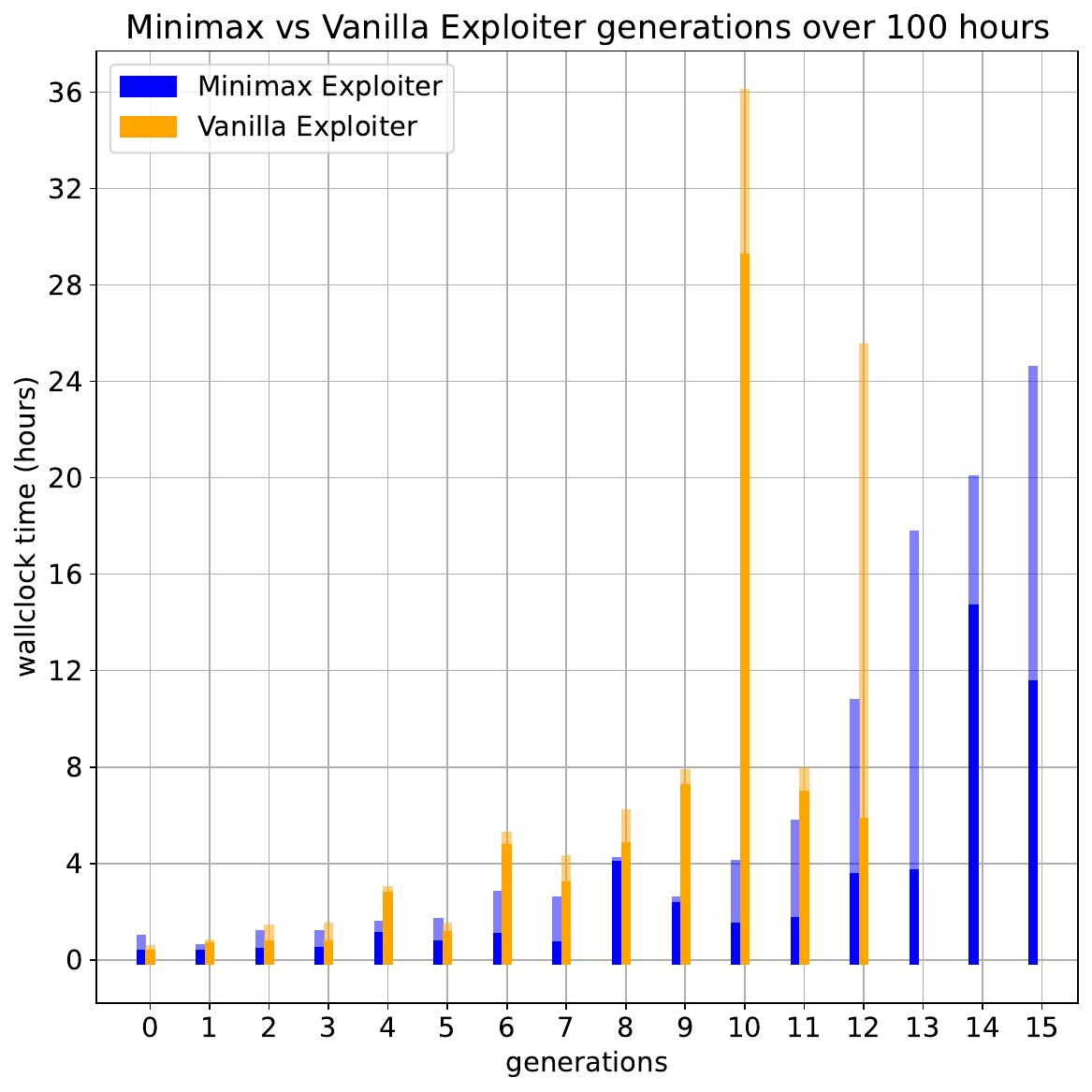}}
\caption{Minimax vs Vanilla Exploiter Performance over $100$ hours in \textsc{For Honor}. The Minimax Exploiter is able to generate $16$ converged Exploiters, while the Vanilla Exploiter only generated $13$ converged Exploiters.}
\label{figure:SmartExploiter_vs_Vanilla_100_performance}
\end{center}
\vskip -0.2in
\end{figure}

We first evaluate the number of converged exploiters within a $24$ hour training period over three seeds. Figure \ref{figure:SmartExploiter_vs_Vanilla_vs_Defensive_vs_Aggressive_performance} shows the results for all four exploiters. We show the results of a longer training session of the Minimax Exploiter and the Vanilla Exploiter for over $100$ hours in Figure~\ref{figure:SmartExploiter_vs_Vanilla_100_performance}. The lower opacity bars in the graphs correspond to the time spent by the Exploiter at convergence. Consistent with previous experiments, we find that the Minimax Exploiter converges faster than the Vanilla Exploiter, and so -- given the same amount of training time -- our system running the Minimax Exploiter can generate more exploiters (counter strategies) into the league.

Finally, we compare the resulting Main Agents generated by the system trained with all four different exploiters, according to their best seed. Each one of them is paired against the other (totalling six pairings) and evaluated over $1000$ duels. We show in Table  \ref{table:main_agents_winrates} that the Main Agent resulting from the Minimax Exploiter experiment is the most robust, achieving a win-rate above $66\%$ against all other Main Agents.

\begin{table*}[t!]
\caption{Win rates of the best final Main Agents generated by all four different Exploiters, averaged over $1000$ duels against each other. The Minimax Main Agent (bold font) is able to beat all other Main Agents, with its best win-rate being $80.26\%$ against the Vanilla Main Agent, and its worse win-rate being $66.25\%$ against the Aggressive Main Agent.}
\label{table:main_agents_winrates}
\vskip 0.15in
\begin{center}
\begin{small}
\begin{sc}
\begin{tabular}{|l|l|l|l|l|}
\hline
 & Vanilla & Minimax & Defensive & Aggressive \\ \hline
Vanilla & \multicolumn{1}{c|}{--} & $19.74\%$ & $66.86\%$ & $24.39\%$ \\ \hline
\textbf{Minimax} & $\textbf{80.26\%}$  & \multicolumn{1}{c|}{--} & $\textbf{70.31\%}$ & $\textbf{66.25\%}$ \\ \hline
Defensive & $33.14\%$ & $29.69\%$ & \multicolumn{1}{c|}{--} & $20.93\%$ \\ \hline
Aggressive & $75.61\%$ & $33.75\%$ & $79.07\%$ & \multicolumn{1}{c|}{--} \\ \hline
\end{tabular}
\end{sc}
\end{small}
\end{center}
\vskip -0.1in
\end{table*}

\section{Discussion}
\label{discussion}
In this paper, we proposed the Minimax Exploiter, a simple to implement change to the league framework that can provide large improvement to training efficiency in certain settings. We argue that the current state of CSP-MARL can be too computationally expensive and time intensive to be run in iterative workflows that require near-constant model retraining, such as video game productions. We showed across several environments, including simple turn-based games, Atari, and a modern AAA video game, that the Minimax Exploiter can achieve gains in overall run-time efficiency. In simple environments we found that simple reward shaping performed well, whereas in a complicated environment like For Honor, we found that the Minimax Exploiter generated much more robust Main Agents (compared to the vanilla, defensive, and aggressive exploiters) in equal amount of training time. As next steps, we plan to further explore the uses of the Minimax Exploiter in asymmetric games. In fact, many games, including For Honor, can be seen as an asymmetric game as there may be a variety of different classes to play as. We hypothesize that the Minimax Exploiter lends itself perfectly to these asymmetric scenarios, as even though the state and action spaces may differ, the evaluation function of the opponent that we proposed in this paper can still be used as a reward for an asymmetric agent. We hope that the flexibility and increased efficiency of our approach makes it more appealing for AAA game productions to run CSP-MARL for their use-cases.




\bibliographystyle{ACM-Reference-Format} 
\bibliography{smartexploiter_references}


\newpage
\appendix
\onecolumn

\section{For Honor RL Interface}
\label{appendix:FH_RL}

We provide more details into the \textsc{For Honor} RL interface. The \textsc{For Honor} RL state consists of a 160 dimensional vector, broken down according to the following sub-vectors:
\begin{itemize}
\item Agent Information: 60 dimensions.
\begin{itemize}
  \item Animation Information: 52 dimensions.
  \item Health: 1 dimension.
  \item Stamina: 1 dimension.
  \item Out of Stamina: 1 dimension.
  \item Stance: 5 dimensions.
\end{itemize}
\item Opponent Information: 60 dimensions.
\begin{itemize}
  \item Animation Information: 52 dimensions.
  \item Health: 1 dimension.
  \item Stamina: 1 dimension.
  \item Out of Stamina: 1 dimension.
  \item Stance: 5 dimensions.
\end{itemize}
\item Target Distance: 1 dimension.
\item Speed: 2 dimensions.
\item Can Parry: 1 dimension.
\item Action Mask: 36 dimensions.
\end{itemize}

We query the RL agent for a decision every 100 milliseconds. Note, however, that this does not mean that the agent \textit{acts} every 100 milliseconds but, rather, that they have an opportunity to input a decision within these timed windows. If the agent is in the middle of an animation/action-sequence, whether it be performing an attack or being the victim of an attack, then our query is rejected and the agent will only have another opportunity to act at the next query (i.e., provided it is free of any of the aforementioned ongoing prohibitive animations).


The RL agent is always operating on what is referred to as the ``lock'' mode, meaning it is always facing its opponent. While the agent has access to a variety of fighting moves, it cannot explicitly control its own movement, with the exception of a few actions -- called ``dodges'' -- which quickly displace the agent along a certain direction. The action space consists of 36 discrete actions (specifically for the Warden game player archetype), broken down into the following categories:
\begin{itemize}
\item Warden Specific Actions: 23 actions.
\begin{itemize}
  \item \{Left, Right, Top\} $\times$ \{Light Attack, Heavy Attack\} : 6 actions.
  \item \{Special Move\} $\times$ \{1 $\xrightarrow{}$ 17\}: 17 actions.
\end {itemize}

\item Generic Actions: 13 actions.
\begin{itemize}
  \item \{Parry, Stance\} $\times$ \{Left, Right, Up\} : 6 actions.
  \item \{Dodge\} $\times$  \{Left, Right, Front, Back\}: 4 actions.
  \item NoOp (do nothing).
  \item Feint.
  \item Guard Break.
\end {itemize}


\end{itemize}

\end{document}